# Large Language Model (LLM) AI text generation detection based on transformer deep learning algorithm

**Yuhong Mo[1*], Hao Qin[2], Yushan Dong[3], Ziyi Zhu[4] , Zhenglin Li[5]**

[1]College of engineering, Carnegie Mellon university, PA, Pittsburgh, 15213, USA
[2]Independent, China
[3]University of Maryland, MD, USA
[4]New York University, USA
[5]Texas A&M University, USA

[*]Corresponding author email: yuhongmo@cmu.edu

**Abstract.** In this paper, a tool for detecting LLM AI text generation is developed based on the Transformer model, aiming to improve the accuracy of AI text generation detection and provide reference for subsequent research. Firstly the text is Unicode normalised, converted to lowercase form, characters other than non-alphabetic characters and punctuation marks are removed by regular expressions, spaces are added around punctuation marks, first and last spaces are removed, consecutive ellipses are replaced with single spaces and the text is connected using the specified delimiter. Next remove non-alphabetic characters and extra whitespace characters, replace multiple consecutive whitespace characters with a single space and again convert to lowercase form. The deep learning model combines layers such as LSTM, Transformer and CNN for text classification or sequence labelling tasks. The training and validation sets show that the model loss decreases from 0.127 to 0.005 and accuracy increases from 94.96 to 99.8, indicating that the model has good detection and classification ability for AI generated text. The test set confusion matrix and accuracy show that the model has 99% prediction accuracy for AI-generated text, with a precision of 0.99, a recall of 1, and an f1 score of 0.99, achieving a very high classification accuracy. Looking forward, it has the prospect of wide application in the field of AI text detection.

**Keywords:** SVM, Stock price prediction, Time series.

## 1. Introduction

AI-generated text detection research refers to the use of artificial intelligence technology to identify and detect generated text content to distinguish real text from false, misleading or inappropriate content [1]. With the rapid development of deep learning technology, the application of AI-generated text is becoming more and more widespread, but at the same time, it also brings some challenges, such as the dissemination of false information, privacy leakage and other problems. Therefore, the detection and identification of AI-generated text has become one of the current hotspots in the field of artificial intelligence [2].

The background of AI-generated text detection research can be traced back to the development of the natural language processing (NLP) field. With the rise of deep learning technology, especially the emergence of models such as Recurrent Neural Networks (RNN) [3], Long Short-Term Memory Networks (LSTM) [4], and Transformer [5], AI has made great progress in text generation. These models can generate high-quality text content such as articles, dialogues, news, etc., but they are also easily used to create false information, deceive users, or spread bad content [6].

Deep learning plays a crucial role in AI generated text. Deep learning models learn linguistic patterns and regularities through training on large amounts of data and are able to generate texts with realism and coherence. Among them, pre-trained language models based on the Transformer architecture (e.g., GPT-3, BERT, etc.) [7] are outstanding in natural language processing tasks, providing powerful support for AI-generated text.

It is due to the wide application and efficient performance of deep learning techniques that there are some problems in AI generated text. For example, false information dissemination, malicious use of AI to generate false news, advertisements and other content to interfere with social opinion; as well as privacy leakage, personal information may be used to generate fraudulent information against specific individuals inappropriate content generation: in addition, inappropriate content involving violence, pornography, hate speech and other inappropriate content may be generated and disseminated through AI.

To address the above issues, there are many researchers working on developing various methods to detect and identify problematic content in AI-generated texts. This includes rule-based, statistical methods, and machine learning techniques (e.g., support vector machines, random forests, etc.) [8] to construct detection models and combining deep learning techniques to improve detection accuracy. In this paper, a tool for detecting LLM AI text generation is developed based on transformer, hoping to improve the accuracy of AI text generation detection and provide some reference and reference for subsequent research.

## 2. Data set sources and data analysis

The data used in this paper comes from the open-source dataset, which includes manually written texts as well as AI-generated texts, where manually written texts are labelled as 0 and AI-generated texts are labelled as 1. Some of the data are shown in Table 1, and in addition, the number of manually written texts and AI-generated texts are counted as shown in Figure 1.

**Table 1.** Some of the data.

| ID | Text | Generated |
|---|---|---|
| 0 | Cars. Cars have been around since they became ... | 0 |
| 1 | Transportation is a large necessity in most co... | 0 |
| 2 | "America's love affair with it's vehicles seem... | 0 |
| 3 | How often do you ride in a car? Do you drive a... | 0 |
| 4 | Cars are a wonderful thing. They are perhaps o... | 0 |

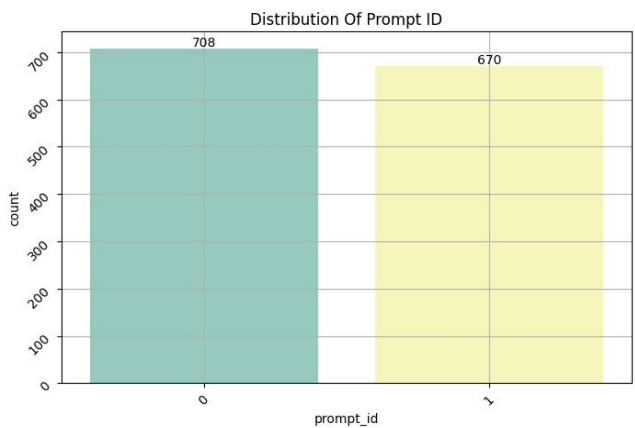

**Figure 1.** The number of manually written texts and AI-generated.
（Photo credit : Original）

As can be seen from Figure 1, there are 708 texts written by humans and 670 texts generated by AI, with roughly the same number of texts for both types.

## 3. Cleaning of this article

The text is first Unicode normalised, then converted to lower case, then regular expressions are used to remove non-alphabetic and punctuation characters, add spaces around punctuation, remove leading and trailing spaces, replace consecutive apostrophes with a single space and finally join the text using the specified separator. Next removes handles starting with @, then removes non-alphabetic characters and extra whitespace, replaces multiple consecutive whitespace characters with a single space, and converts the text to lowercase form.

In this paper, we use python to define a TextVectorization layer that uses a custom Clean function to normalise the text data, set parameters such as maximum number of features, range of ngrams, output mode and sequence length and configure the layer by adapting it to the training data to finally convert the text in the training data to a vectorial representation in the form of integers.

## 4. Method

The deep learning model used in this paper combines different types of layers such as LSTM, Transformer and CNN for tasks such as text classification or sequence annotation. The structure of the model is shown in Fig. 2 and the specific parameters are shown in Table 2.

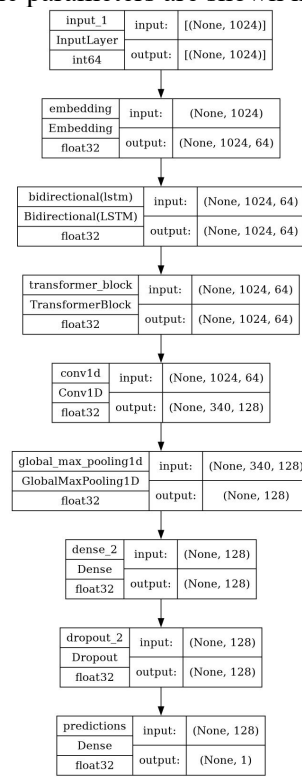

**Figure 2.** The structure of the model.
（Photo credit : Original）

Firstly, the input goes through an Embedding layer to convert the text sequence into a dense vector representation, and then the sequence features are extracted by a bidirectional LSTM. Then, a custom TransformerBlock module is introduced, which contains the multi-head attention mechanism and feed-forward neural network part [9]. In TransformerBlock, global dependencies are captured by the multi-head attention mechanism and feature transformation and nonlinear mapping are performed by the feed-forward neural network.

In the model structure, TransformerBlock is applied on the sequence representation of the LSTM output to improve the model's ability to model long distance dependencies. The stacking and residual concatenation mechanism through the TransformerBlock helps to mitigate the gradient vanishing problem and allows the model to better capture the complex features of the input sequences. Afterwards, a one-dimensional convolutional layer (Conv1D) is applied on the TransformerBlock output for further extracting local features and reducing the sequence length. Next, a global maximum pooling layer (GlobalMaxPooling1D) is used to fuse the features at different locations into a fixed-length vector representation.

This is followed by a fully connected layer (Dense) and a Dropout layer for learning higher level feature representations and preventing overfitting. The last layer is the Dense layer with Sigmoid activation function for probabilistic output in binary classification tasks. The whole model is constructed using a functional API, where inputs and outputs are defined in the Model class to form an end-to-end trainable deep learning model [10].

The structure of this model incorporates different types of neural network layers and makes full use of the respective advantages of structures such as LSTM, Transformer and CNN when dealing with sequence data, aiming to improve the performance of the model when classifying or labelling text data. At the same time, technical means such as residual connection, multi-head attention mechanism and convolutional operation are used to make the model have better modelling ability and generalisation ability, and perform well when dealing with tasks such as text classification.

Table 2. The specific parameters.

| Layer(type) | Output Shape | Param# |
| --- | --- | --- |
| input 1(InputLayer) | [(None,1024)] | 0 |
| embedding(Embedding) | (None,1024,64) | 4800000 |
| bidirectional(Bidirectional) | (None,1024,64) | 24832 |
| transformer block(TransformerBlock) | (None,1024,64) | 37664 |
| conv1d(Conv1D) | (None,340,128) | 57472 |
| global max pooling1d(GlobalMaxPooling1D) | (None,128) | 0 |
| dense 2(Dense) | (None,128) | 16512 |
| dropout 2(Dropout) | (None,128) | 0 |
| predictions(Dense) | (None,1) | 129 |

## 5. Result

In the experimental setup, two callback functions are defined, the first one is ModelCheckpoint, which is used to save the weights of the model with the best performance on the validation set to the file "model.h5" and only the best model is saved; the second one is EarlyStopping, which is used to stop the training in advance when the monitoring metrics on the validation set are no longer improving and restore the best model weights to the file "model.h5". Weights. The models were then compiled using the Adam optimiser, binary cross-entropy loss function and accuracy as evaluation metrics. Next, training is performed using x_train and y_train data, 10 epochs are set, 10% of the training data is used as the validation set for verification, and the two callback functions defined earlier are passed in via the callbacks parameter to achieve model saving and early stopping functions during model training. The whole process aims to train a deep learning model with good performance and generalisation ability. The variations of loss and accuracy for the first 4 epoch model training set and validation set are shown in Table 3. The model is tested using the test set and the output confusion matrix is shown in Fig. 3 and the evaluation metrics of the output model are shown in Table 4.

Table 3. The variations of loss and accuracy.

| Epoch | Train loss | Train accuracy | Val loss | Val accuracy |
| --- | --- | --- | --- | --- |
| 0 | 0.127258 | 0.949625 | 0.019851 | 0.991674 |
| 1 | 0.008398 | 0.997121 | 0.040408 | 0.987974 |
| 2 | 0.00605 | 0.998149 | 0.046024 | 0.990749 |
| 3 | 0.005541 | 0.998972 | 0.021248 | 0.990749 |

From the changes of loss and accuracy in the training and validation sets, the value of loss gradually decreases from 0.127 to 0.005, and the prediction accuracy of the model gradually increases from 94.96 to 99.8, which indicates that the model built in this paper is able to detect and classify AI-generated text well.

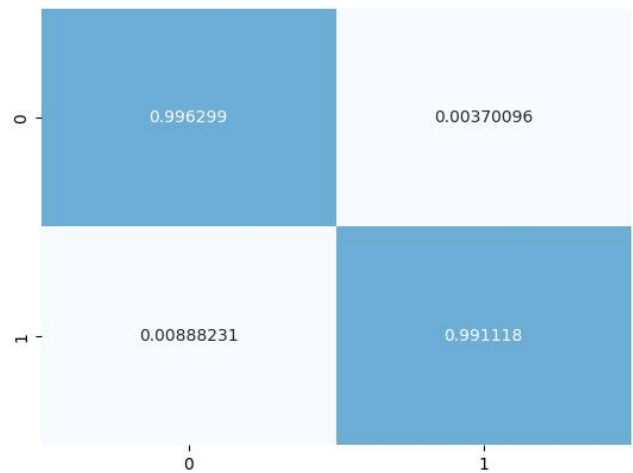

**Figure 3.** Confusion matrix.
（Photo credit : Original）

**Table 4.** The evaluation metrics of the output model.

| | Precision | Recall | F1-score | Support |
|---|---|---|---|---|
| | 0.99 | 1 | 0.99 | 1351 |
| | 1 | 0.99 | 0.99 | 1351 |
| Accuracy | | | 0.99 | 2702 |
| Macro avg | 0.99 | 0.99 | 0.99 | 2702 |
| Weighted avg | 0.99 | 0.99 | 0.99 | 2702 |

As shown by the confusion matrix and accuracy of the test set, the model achieves 99% accuracy in the prediction of AI-generated text, in addition to a precision of 0.99, a recall of 1, and an f1 score of 0.99, which achieves a very high classification accuracy, and hopefully can be applied to the field of AI paper detection in the future.

## 6. Conclusion
The LLM AI text generation detection tool developed based on Transformer in this paper has achieved significant results in improving the accuracy of AI text generation detection and provides important reference and reference value for subsequent research.
Firstly, the text data is effectively cleaned through operations such as Unicode normalisation of the text, conversion to lowercase form, and regular expression removal of characters other than non-alphabets and punctuation marks. When dealing with punctuation, the steps of adding spaces, removing first and last spaces, and replacing consecutive ellipses with single spaces make the text more normalised, which is conducive to the accuracy of subsequent model training and prediction.

Secondly, different types of layers such as LSTM, Transformer and CNN are combined in the model design for text classification tasks. By observing the changes of loss as well as accuracy in the training and validation sets, it can be seen that the model is gradually optimised during the training process, and the loss value decreases from 0.127 to 0.005, while the accuracy increases from 94.96% to 99.8%, which indicates that the model is able to effectively detect and classify AI-generated text.

Further, the confusion matrix and accuracy analysis on the test set shows that the model achieves 99% prediction accuracy for AI-generated text with a precision of 0.99, a recall of 1, and an f1 score of 0.99, demonstrating a very high classification performance. These results indicate that the proposed model has high reliability and accuracy in the field of AI-generated text detection.

Overall, through the method and model proposed in this study, satisfactory results have been achieved in the area of AI-generated text detection, and it has a strong potential for popularisation and application. It is hoped that the model can be further improved in the future and widely applied in the field of AI-generated text detection to make a greater contribution to guaranteeing information security and content authenticity.